\documentclass[p,twocolumn,twoside,showpacs,byrevtex,superscriptaddress]{revtex4}

\usepackage{currfile}

\usepackage{graphicx,epsfig,verbatim,enumerate}
\usepackage{amssymb,amsmath}
\usepackage{ifthen}
\newboolean{twocolswitch}

\usepackage{placeins}

\newcommand{\sindex}[1]{}
\newcommand{\nindex}[1]{}

\newcommand{\www}[1]{\url{#1}}

\usepackage{lettrine}

\usepackage{color}

\setboolean{twocolswitch}{true}

\begin{document}

\title{
  
  Sifting Robotic from Organic Text:  A Natural Language Approach for Detecting Automation on Twitter

}

\author{
\firstname{Eric M.}
\surname{Clark}
}

\email{eclark@uvm.edu}

\affiliation{Department of Mathematics \&  Statistics}
\affiliation{Vermont Complex Systems Center}
\affiliation{Vermont Advanced Computing Core}
\affiliation{Computational Story Lab}
\affiliation{Department of Surgery}

\author{
\firstname{Jake Ryland}
\surname{Williams}
}

\affiliation{Department of Mathematics \&  Statistics}
\affiliation{Vermont Complex Systems Center}
\affiliation{Vermont Advanced Computing Core}
\affiliation{Computational Story Lab}

\author{
\firstname{Chris A.}
\surname{Jones}
}

\affiliation{Department of Surgery}
\affiliation{Global Health Economics Unit of the Vermont Center for Clinical and Translational Science}
\affiliation{Vermont Center for Behavior and Health}

\author{
\firstname{Richard A.}
\surname{Galbraith}
}

\affiliation{Department of Medicine}
\affiliation{Vermont Center for Clinical and Translational Science}

\author{
\firstname{Christopher M.}
\surname{Danforth}
}

\affiliation{Department of Mathematics \&  Statistics}
\affiliation{Vermont Complex Systems Center}
\affiliation{Vermont Advanced Computing Core}
\affiliation{Computational Story Lab}

\author{
\firstname{Peter Sheridan}
\surname{Dodds}
}

\affiliation{Department of Mathematics \&  Statistics}
\affiliation{Vermont Complex Systems Center}
\affiliation{Vermont Advanced Computing Core}
\affiliation{Computational Story Lab}

\date{\today}

\begin{abstract}
  
Twitter, a popular social media outlet, has evolved into a vast source of linguistic data, rich with opinion, sentiment, and discussion.  Due to the increasing popularity of Twitter, its perceived potential for exerting social influence has led to the rise of a diverse community of automatons, commonly referred to as bots.  These inorganic and semi-organic Twitter entities can range from the benevolent (e.g., weather-update bots, help-wanted-alert bots) to the malevolent (e.g., spamming messages, advertisements, or radical opinions). Existing detection algorithms typically leverage metadata (time between tweets, number of followers, etc.) to identify robotic accounts. Here, we present a powerful classification scheme that exclusively uses the natural language text from organic users to provide a criterion for identifying accounts posting automated messages.  Since the classifier operates on text alone, it is flexible and may be applied to any textual data beyond the Twittersphere.
  
\end{abstract}

\maketitle

\section{Introduction}
Twitter has become a mainstream social outlet for the discussion of a myriad of topics through microblogging interactions.   Members chiefly communicate via short text-based public messages restricted to 140 characters, called tweets.  As Twitter has evolved from a simple microblogging social media interface into a mainstream source of communication for the discussion of current events, politics, consumer goods/services, it has become increasingly enticing for parties to gameify the system by creating automated software to send messages  to organic (human) accounts as a means for personal gain and for influence manipulation \cite{subrahmanian2016darpa,EvilDataSci}.  The results of sentiment and topical analyses can be skewed by robotic accounts that dilute legitimate public opinion by algorithmically generating vast amounts of inorganic content.
Nevertheless, data from Twitter is becoming a source of interest in public health and economic research in monitoring the spread of disease \cite{sadilek2012modeling,wagstaff2012four}  and  gaining insight into public health trends \cite{mitchell2013geography}.  
 
 \par
  
 In related work \cite{BotorNot,DetectingSpammersTwitter,HumanBotCyborgTwitter,DetectingAutomationTwitter}, researchers have built classification algorithms using metadata idiosyncratic to Twitter, including the number of followers, posting frequency, account age, number of user mentions/replies, username length, and number of retweets.  However, relying on metadata can be problematic: sophisticated spam algorithms now emulate the daily cycle of human activity and author borrowed content to appear human \cite{BotorNot}.    Another problematic spam tactic is the renting of accounts of legitimate users (called sponsored accounts), to introduce short bursts of spam and hide under the user's organic metadata to mask the attack \cite{SuspendedRetroTwitterSpam}.  
 
 \par
 A content based classifier proposed by \cite{chu2012detecting} measures the entropy between Twitter time intervals along with user meta data to classify Twitter accounts,  and requires a comparable number of tweets ($\geq 60$) for adequate classification accuracy as our proposed method.  SentiBot, another content based classifier \cite{dickerson2014using}, utilizes  latent Dirichlet allocation (LDA) for topical categorization combined with sentiment analysis techniques to classify individuals as either bots or humans.  We note that as these automated entities evolve their strategies, combinations of our proposed methods and studies previously mentioned may be required to achieve reasonable standards for classification accuracy.  Our method classifies accounts solely based upon their linguistic attributes and hence can easily be integrated into these other proposed strategies.  
    
\par
We introduce a classification algorithm that operates using three linguistic attributes of a user's text.  The algorithm analyzes: \begin{enumerate}
\item the average URL count per tweet
\item the average pairwise lexical dissimilarity between a user's tweets, and
\item the word introduction rate decay parameter of the user for various proportions of time-ordered tweets
\end{enumerate}

We provide detailed descriptions of each attribute in the next section.  We then test and validate our algorithm  on 1 000 accounts which were hand coded as automated or human. 
\par
  We find that for organic users, these three attributes are densely clustered, but can vary greatly for automatons.  We compute the average and standard deviation of each of these dimensions for various numbers of tweets from the human coded organic users in the dataset.  We classify accounts by their distance from the averages from each of these attributes.  The accuracy of the classifier increases with the number of tweets collected per user.  Since this algorithm operates independently from user metadata, robotic accounts do not have the ability to adaptively conceal their identities by manipulating their user attributes algorithmically.  Also, since the classifier is built from time ordered tweets, it can determine if a once legitimate user begins demonstrating dubious behavior and spam tactics.  This allows for social media data-miners to dampen a noisy dataset by weeding out suspicious accounts and focus on purely organic tweets.

\section{Data Handling}
\subsection{Data-Collection}

   We filtered a 1\% sample of Twitter's streaming API (the spritzer feed) for tweets containing geo-spatial metadata spanning the months of April through July in 2014.  Since roughly $1\%$ of tweets provided GPS located spatial coordinates, our sample represents nearly all of the tweets from users who enable geotagging.  This allows for much more complete coverage of each user's account.  From this sample, we collected all of the geo-tweets from the most active 1 000 users for classification as human or robot and call this the Geo-Tweet dataset.  

\subsection{Social HoneyPots}

 To place our classifier in the context of recent work, we applied our algorithm to another set of accounts collected from the Social HoneyPot Experiment  \cite{7MonthsDevils}.  This work exacted a more elaborate approach to find automated accounts on Twitter by creating a network of fake accounts (called Devils \cite{DevilsAngelsRobots}) that would tweet about trending topics amongst themselves in order to tempt robotic interactions.  The experiment was analyzed and compiled into a dataset containing the tweets of ``legitimate users" and those classified as ``content polluters".   We note that the users in this dataset were not hand coded.  Accounts that followed the Devil honeypot accounts were deemed robots.  Their organic users were compiled from a random sample of Twitter, and were only deemed organic because these accounts were not suspended by Twitter at the time.  Hence the full HoneyPot dataset can only serve as an estimate of the capability of this classification scheme.

 \subsection{Human Classification of Geo-Tweets}
 
   Each of the 1 000 users were hand classified separately by two evaluators.  All collected tweets from each user were reviewed until the evaluator noticed the presence of automation.  If no subsample of tweets appeared to be algorithmically generated, the user was classified as human.  The results were merged, and conflicting entries were resolved to produce a final list of user ids and codings.  See Figure 1 for histograms and violin plots summarizing the distributions of each user class.  We note that any form of perceived automation was sufficient to deem the account as automated.  See SI for samples of each of these types of tweets from each user class and a more thorough description of the annotation process. 
     
  \begin{figure}[tbp!]
 \centering
 \includegraphics[width=0.99\columnwidth]{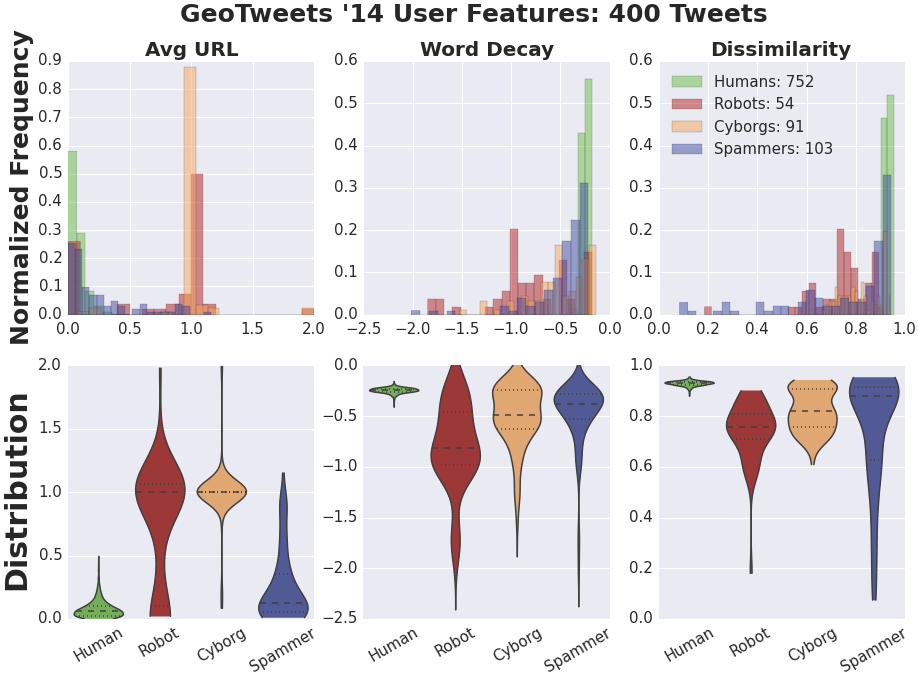}
 \caption{ The feature distribution of the 1000 hand coded users are summarized with histograms and violin plots.  These show the wide variation in automated features versus Organics.  Violin plots show the kernel density estimation of each distribution.  Using the Organic features, automated entities are identified by exclusion.}
\label{fig:UserFeature}
 \end{figure}

 \subsection{ Types of Users}
    
 We consider organic content, i.e. from human accounts, as those that have not tweeted in an  
 algorithmic fashion.  We focused on three distinct classes of automated tweeting: \\
 
 \textbf{Robots:}  Tweets from these accounts draw on a strictly limited vocabulary.  The messages follow a very structured pattern, many of which are in the form of automated updates. Examples include Weather Condition Update Accounts, Police Scanner Update Accounts, Help Wanted Update Accounts, etc.  \\
 
 \textbf{Cyborgs:}  The most covert of the three, these automatons exhibit human-like behavior and messages through loosely structured, generic, automated messages and from borrowed content copied from other sources.   Since many malicious cyborgs on Twitter try to market an idea or product, a high proportion of their tweets contain URLs, analogous to spam campaigns studied on Facebook \cite{SocialSpamCampaigns}.  Messages range from the backdoor advertising of goods and services \cite{xExamEcigTwiter} to those trying to influence social opinion or even censor political conversations \cite{SpamPoliticalCensorship}.  These accounts act like puppets from a central algorithmic puppeteer to push their product on organic users while trying to appear like an organic user \cite{MarionetteMicrobloggers}.  Since these accounts tend to borrow content, they have a much larger vocabulary in comparison to ordinary robots.  Due to Twitter's 140 character-per-tweet restriction, some of the borrowed content being posted must be truncated.  A notable attribute of many cyborgs is the presence of incomplete messages followed by an ellipsis and a URL.  Included in this category are `malicious promoter' accounts \cite{7MonthsDevils} that are radically promoting a business or an idea systematically.  \\
 
 \textbf{Human Spammers:}   These are legitimate accounts that abuse an algorithm to post a burst of almost indistinguishable tweets that may differ by a character in order to fool Twitter's spam detection protocols.  These messages are directed at a particular user, commonly for a follow request to attempt to increase their social reach and influence.    
 
   Although we restrict our focus to the aforementioned classes, we did notice the presence of other subclasses, which we have named ``listers", and ``quoters", that have both organic and automaton features.  Listers are accounts that send their messages to large groups of individuals at once.  Quoters are dedicated accounts that are referencing distant passages from literature or song lyrics.  Most of the tweets from these accounts are all encased in quotations.  These accounts also separately tweet organic content.  We classified these accounts as human because there was not sufficient evidence suggesting these behaviors were indeed automated. 
  \section{Methods}
 
 \subsection{Classification Algorithm}
 
    The classifier, $\mathcal{C}$,  takes ordinal samples of tweets from each user, $\mu$, of varying number, $s$, to determine if the user is a human posting strictly organic content or is algorithmically automating tweets:
    $$ \mathcal{C}: \mu^s \rightarrow \{0,1\} = \{ \text{Organic, Automaton} \}.$$
  Although we have classified each automaton into three distinct classes, the classifier is built more simply to detect and separate organic content from automated.  To classify the tweets from a user, we measure three distinct linguistic attributes: 
  \begin{enumerate}
  \item Average Pairwise Tweet Dissimilarity, 
  \item  Word Introduction Rate Decay Parameter, 
  \item  Average number of URLs (hyperlinks) per tweet.
  \end{enumerate}

\subsection{Average Pairwise Tweet \\ Dissimilarity}

   Many algorithmically generated tweets contain similar structures with minor character replacements and long chains of common substrings.  Purely organic accounts have tweets that are very dissimilar on average.  The length of a tweet, $t$, is defined as the number of characters in the tweet and is denoted $|t|$.  Each tweet is cleaned by truncating multiple whitespace characters and the metric is performed case insensitively.  A sample of $s$ tweets from a particular user is denoted  $T^s_\mu$. Given a pair of tweets from a particular user, $t_i,t_j\in T^s_\mu$, the pairwise tweet dissimilarity, $D(t_i,t_j)$, is given by subtracting the length of the longest common subsequence of both tweets, $|LCS(t_i,t_j)|$ and then weighting by the sum of the lengths of both tweets:
   $$ \text{D}(t_i,t_j) = \dfrac{ |t_i| + |t_j| - 2\cdot|LCS(t_i,t_j)|}{ |t_i|+|t_j|}. $$
   The average tweet dissimilarity of user $\mu$ for sample size of $s$ tweets is calculated as:
   
   $$\mu_{lcs}^s = \frac{1}{{s \choose 2}} \cdot  \sum_{t_i,t_j \in T^s_\mu} D(t_i,t_j)  . $$
   
   \par
   
   For example, given the two tweets: \\ $(t_1, t_2)$ = (I love Twitter, I love to spam).  Then $|t_1|$ = $|t_2|$ = 14, $LCS(t_1,t_2) = |\text{I love t}| = 8$ (including whitespaces) and we calculate the pairwise tweet dissimilarity as: 
   $$D(t_1,t_2) = \dfrac{14 + 14 - 2\cdot 8 }{14+14} = \dfrac{12}{28} = \dfrac{3}{7}.   $$
 
\subsection{Word Introduction Decay Rate}
  Since social robots \emph{automate} messages, they have a limited and crystalline vocabulary in comparison to organic accounts.  Even cyborgs that mask their automations with  borrowed content cannot fully mimic the rate at which organic users introduce unique words into their text over time.  The word introduction rate is a measure of the number of unique word types introduced over time from a given sample of text \cite{TxtMixing}.  The rate at which unique words are introduced naturally decays over time, and is observably different between automated and organic text.  By testing many random word shufflings of a text, we define $\overline{m_n}$ as the average number of words between the $n^{th}$ and $n+1^{st}$ initial unique word type appearances.  From \cite{TxtMixing}, the word introduction decay rate, $\alpha(n)$,  is given as 
  $$\alpha(n) = 1/\overline{m_n} \propto n^{-\gamma} \text{  \hspace{5mm} for $\gamma>0$}.$$
   For each user, the scaling exponent of the word introduction decay rate, $\alpha$, is approximated by performing standard linear regression on the  last third of the log-transformed tail of the average gap size distribution as a function of word introduction number, $n$ \cite{TxtMixing}.  In figure \ref{fig:WI_feats} below,  the log transformed rank-unique word gap distribution is given for each individual in the data set.  Here the human population (green) is distinctly distributed in comparison to the automatons. 
 
 \begin{figure}[Ht]
 \centering
  \includegraphics[width=0.99\columnwidth]{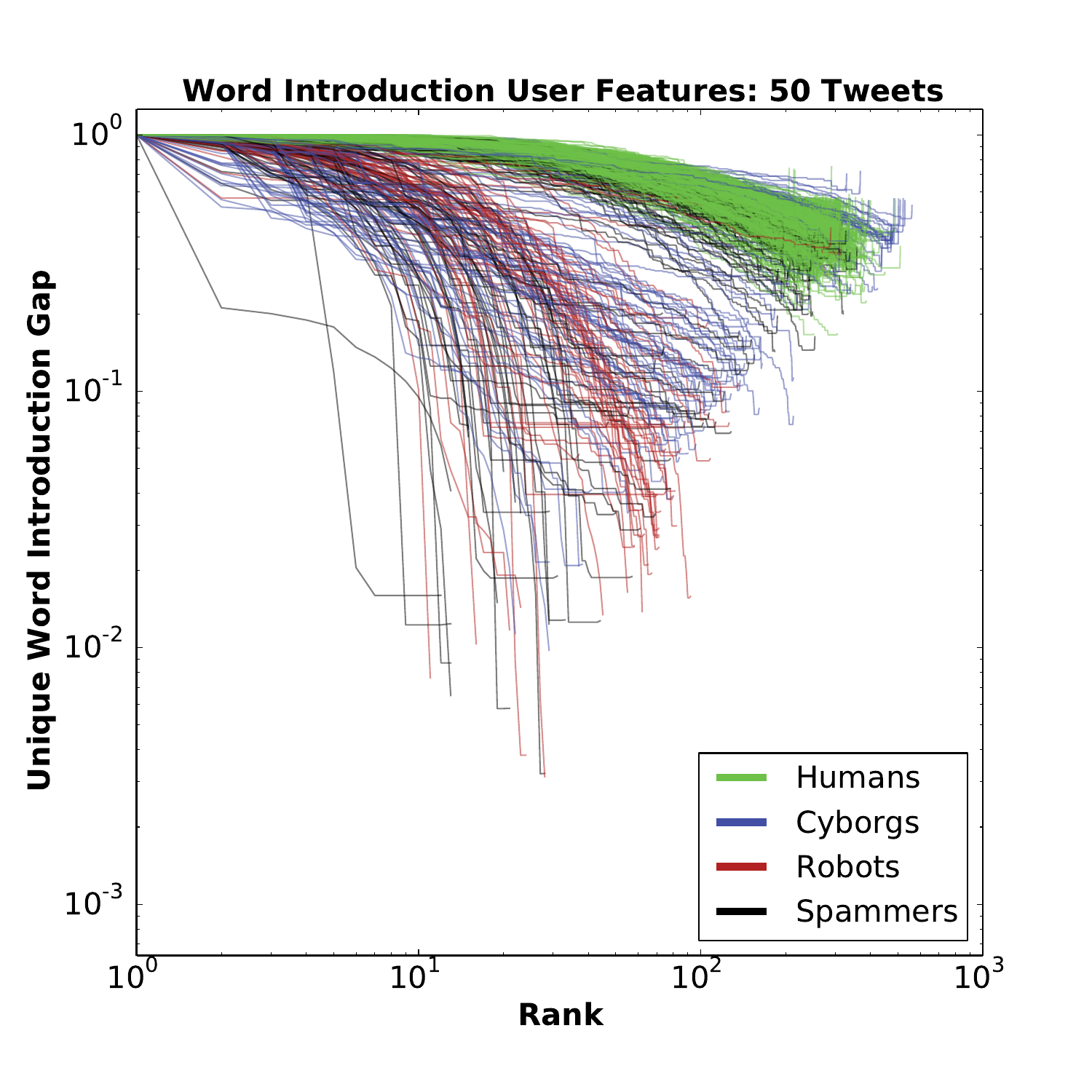}
\caption{\small The rank-unique word gap distribution is plotted on a logscale for each user class.}
\label{fig:WI_feats}
 \end{figure}

\subsection{ Average URLs per Tweet}    

	Hyperlinks (URLs) help automatons spread spam and malware \cite{SuspendedRetroTwitterSpam,ContextAwareSpam,Wagner_whensocial} .  A high fraction of tweets from spammers tend to contain some type of URL in comparison to organic individuals, making the average URLs per tweet a valuable attribute for bot classification algorithms \cite{ HumanBotCyborgTwitter, SocialHoneyPotProtecting, SocialHoneypotsMachineLearning}.  For each user, the average URL rate is measured by the total number of occurrences of the substring `http:' within tweets, and then divided by the total number of tweets authored by the user in the sample of size $s$:  
	$$\mu_{url}^s   = \dfrac{ \#\text{Occurrences of `http:'}}{\#\text{Sampled Tweets}}. $$

 \subsection{Cross Validation Experiment}
   We perform a standard 10-fold Cross Validation procedure on the 2014 Geo-Tweet data set to measure the accuracy of using each linguistic feature for classifying Organic accounts.  We divided individuals  into 10 equally sized groups.  Then 10 trials are performed where 9 of the 10 groups are used to train the algorithm to classify the final group.  
   
   \par 
      During the Calibration phase, we measure each of the three features for every human coded account in the training set.  We sequentially collect tweets from each user from a random starting position in time.  We record the arithmetic mean  and standard deviation of the Organic attributes to classify the remaining group.  The classifier distinguishes Human from Automaton by using a varying threshold, $n$, from the average attribute value computed from the training set.  For each attribute, we classify each user as an automaton if their feature falls further than $n$ standard deviations away from the organic mean, for varying $n$.  
   
   \par
      
      For each trial, the False Positives and True Positives for a varying window size, $n$, are recorded.  To compare to other bot-detection strategies, we rate True Positives as the success at which the classifier identifies automatons by exclusion, and False Positives as humans that are incorrectly classified as automatons.  The results of the trials for varying tweet sizes are averaged and visualized with a Receiver Operator Characteristic curve (ROC) (see Figure \ref{fig:roc_geo}).  The accuracy of each experiment is measured as the area under the ROC, or AUC.  To benchmark the classifier, a 10-fold cross validation was also performed on the HoneyPot  tweet-set which we describe in the following section. \\

 \section{Results and Discussion}
 
 \subsection{Geo-Tweet Classification Validation}
  
\begin{figure}[Ht]
 \centering
  \includegraphics[width=0.99\columnwidth]{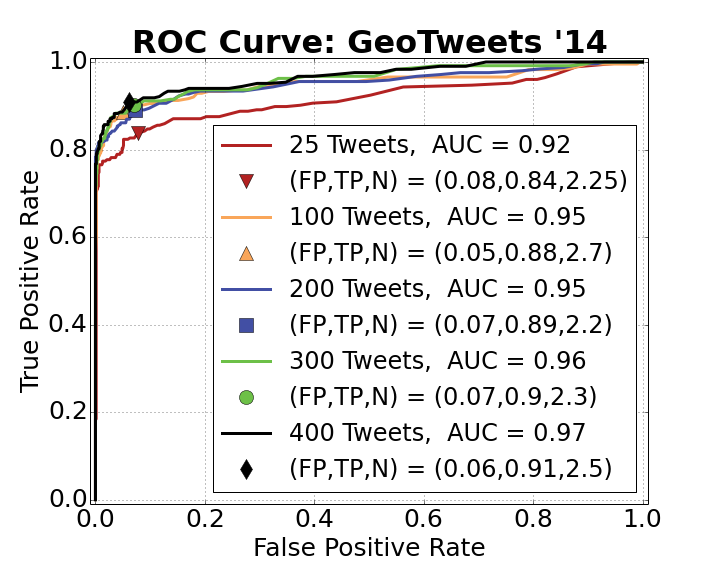}
\caption{\small The receiver operator characteristic curve from the 10-fold Cross Validation Experiment performed on the Geo Tweets collected from April through July 2014.  The True Positive (TP), False Positive (FP), and thresholds, $N$, are averaged across the 10 trials. The accuracies are approximated by the AUCs, which we compute using the trapezoid rule.   The points depict the best experimental model thresholding window (N). }
\label{fig:roc_geo}
 \end{figure}

   The ROC curves for the Geo-Tweet 10 fold Cross Validation Experiment  for varying tweet bins  in Figure \ref{fig:roc_geo} show that the accuracy increases as a function of number of tweets.  

 \begin{figure}[Ht]
 \centering
 \includegraphics[width=0.99\columnwidth]{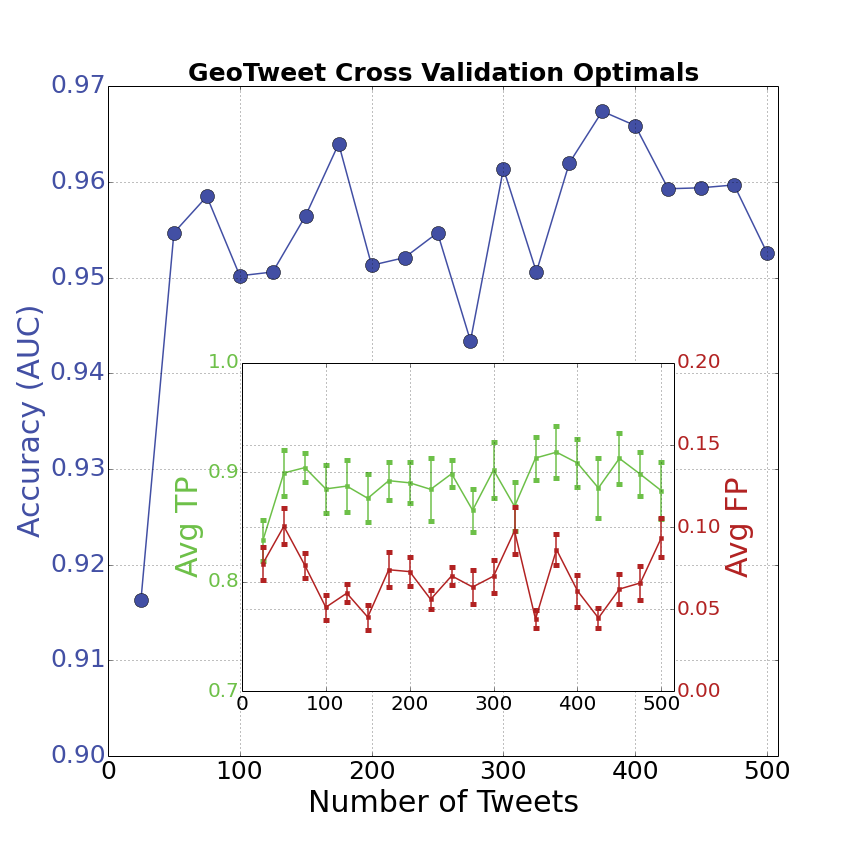}
 \caption{\small Accuracy, computed as the AUC is plotted as a function of number of tweets, ranging from 25 to 500.  The average True Positive and False Positive Rates over 10 trials is given on twin axes with error bars drawn using the standard error.  }
\label{fig:GeoValOpts}
 \end{figure}
 
 \par
 
  Although the accuracy of the classifier  increases with the number of collected tweets, we see in Figure \ref{fig:GeoValOpts} that within 50 tweets the accuracy of the average of 10 random trials is only slightly higher than a 500 tweet user sample.  While this is very beneficial to our task (isolating humans), we note that larger samples see greater returns when one instead wants to isolate spammers, that tweet random bursts of automation.

 \subsection{ HoneyPot External Validation }
 \par
\begin{figure}[Ht]
 \centering
  \includegraphics[width=0.99\columnwidth]{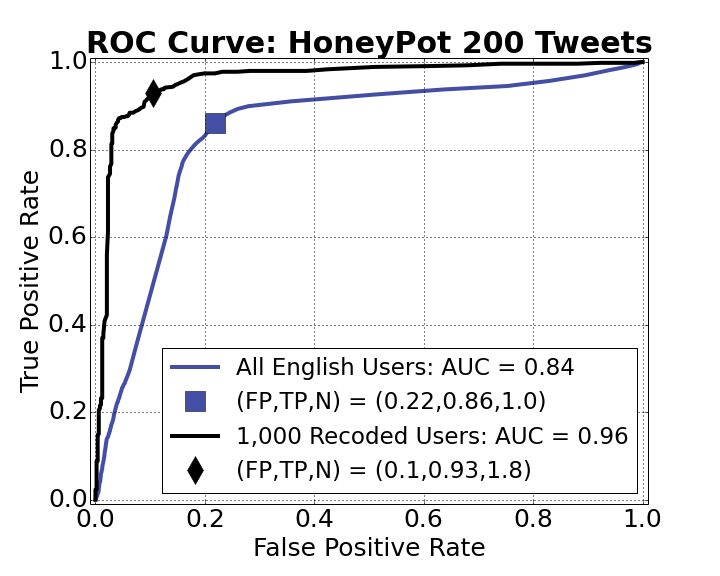}
 \caption{ \small Honey Pot Data Set, 10 fold Cross Validation Performance for users with 200 tweets.  The black curve represents the 1 000 hand coded HoneyPot users, while the blue curve is the entire English Honeypot dataset. The accuracy increases from $84\%$ to $96\%$.   }
\label{fig:roc_hpot}
 \end{figure}

  The classifier was tested on the Social Honeypot Twitter-bot dataset provided by \cite{7MonthsDevils}.  Results are visualized with a ROC curve in Figure \ref{fig:roc_hpot}.  The averaged optimal threshold for the full English user dataset (blue curve) had a high true positive rate (correctly classified automatons: 86\%), but also had a large false positive rate (misclassified humans: 22\%).  
  
  \par The Honeypot Dataset relied on Twitter's spam detection protocols to label their randomly collected ``legitimate users".  Some forms of automation (weather-bots, help-wanted bots) are permitted by Twitter.   Other cyborgs that are posting borrowed organic content can fool Twitter's automation criterion. This ill formation of the training set greatly reduces the ability of the classifier to distinguish humans from automatons, since the classifier gets the wrong information about what constitutes a human.  To see this, a random sample of 1 000  English Honeypot users was hand-coded to mirror the previous experiment.  On this smaller sample (black curve in Figure 4), the averaged optimal threshold accuracy increased to 96\%.  
    
 \subsection{ Calibrated Classifier Performance}
\begin{figure}[Ht]
 \centering
    \includegraphics[width=0.99\columnwidth]{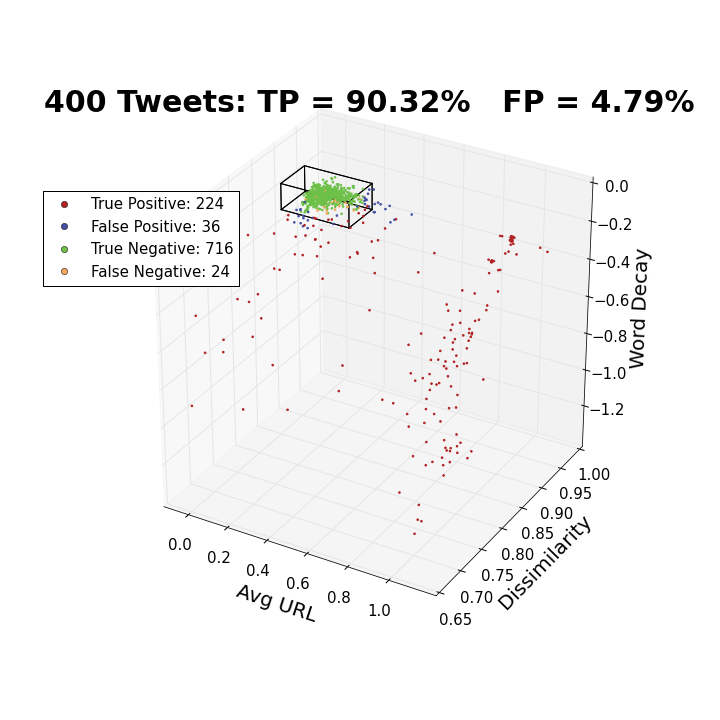}                        
 \caption{\small Calibrated Classifier Performance on 1 000 User Geo-Tweet Dataset.  Correctly classified humans (True Negative), are coded in Green, while correctly identified automatons (True Positives) are coded in red.  The 400 tweet average optimal thresholds from the cross validation experiment designate the thresholding for each feature.  The black lines demonstrates each feature cutoff.}
\label{fig:robotfence}
 \end{figure}

   We created the thresholding window of final calibrated classifier using the results from the calibration experiment.  We average the optimal parameters from the 10 fold cross validation on the Geo-Tweet dataset from each of the 10 calibration trials for tweet bins ranging from 25 to 500 in increments of 25 tweets.  We also average and record the optimal parameter windows, $n_{opt}$ and their standard deviations, $\sigma_{opt}$.  The standard deviations serve as a tuning parameter to increase the sensitivity of the classifier, by increasing the feature cutoff window ($n$). The  results from applying the calibrated classifier to the full set of 1 000 users, using 400 tweet bags is given in Figure \ref{fig:robotfence}.  The feature cutoff window (black lines) estimates if the user's content is organic or automated.  Human feature sets (True Negatives: 716) are densely distributed with a 4.79\% False Positive Rate (i.e., humans classified as robots).   The classifier accurately classified 90.32\% of the automated accounts and 95.21\% of the Organic accounts.  See Figure S1 for a cross sectional comparison of each feature set.  We note that future work may apply different methods in statistical classification to optimize these feature sets, and that  using these simple cutoffs already leads to a high level of accuracy.

  \section{Conclusion}
    Using a flexible and transparent classification scheme, we have demonstrated the potential of using linguistic features as a means of classifying automated activity on Twitter.  Since these features do not use the metadata provided by Twitter, our classification scheme may be applicable outside of the Twittersphere.  Future work can extend this analysis multilingually and incorporate additional feature sets with an analogous classification scheme.  URL content can also be more deeply analyzed to identify organic versus SPAM related hyperlinks. 
      
 \par   
 
 We note the potential for future research to investigate and to distinguish between each sub-class of automaton. We formed our taxonomy according to the different modes of text production.  Our efforts were primarily focused in separating any form of automation from organic,human content.  In doing so we recognized three distinct classes of these types of automated accounts.  However, boundary cases (e.g.  cyborg-spammers, robot-spammers, robotic-cyborgs, etc.) along with other potential aforementioned subclasses (e.g. listers, quoters, etc.) can limit the prowess of our current classification scheme tailored towards these subclasses.  We have shown that human content is distinctly different from these forms of automation, and that for a binary classification of automated or human, these features have a very reasonable performance with 
our proposed algorithm.  

    Our study distinguishes itself by focusing on automated behavior that is tolerated by Twitter, since both types of inorganic content can skew the results of sociolinguistic analyses.  This is particularly important, since Twitter has become a possible outlet for health economics \cite{wagstaff2012four} research including monitoring patient satisfaction and modeling disease spread \cite{broniatowski2013national,sadilek2012modeling}.   Monitoring excessive social media marketing of electronic nicotine delivery systems (also known as e-cigarettes), discussed in \cite{clark2015vaporous, huang2014cross}, makes classifying organic and automated activity relevant for research that can benefit policy-makers regarding public health agendas.  Isolating  organic content on Twitter can help dampen noisy data-sets and is pertinent for research involving social media data and other linguistic data sources where a mixture of humans and automatons exist. \\
   \par
    In health care, a cardinal problem with the use of electronic medical records is their lack of interoperability.  This is compounded by a lack of standardization and use of data dictionaries which results in a lack of precision concerning our ability to collate signs, symptoms, and diagnoses.   The use of millions or billions of tweets concerning a given symptom or diagnosis might help to improve that precision.  But it would be a major setback if the insertion of data tweeted from automatons would obscure useful interpretation of such data.  We hope that the approaches we have outlined in the present manuscript will help alleviate such problems.

\section{Acknowledgments}
The authors wish to acknowledge the Vermont Advanced Computing Core which provided High Performance Computing resources contributing to the research results. EMC and JRW was supported by the UVM Complex Systems Center, PSD was supported by NSF Career Award \# 0846668. CMD and PSD were also supported by a grant from the MITRE Corporation and NSF grant \#1447634.  CJ is supported in part by the National Institute of Health (NIH) Research wards R01DA014028 \& R01HD075669, and by the Center of Biomedical Research Excellence Award P20GM103644 from the National Institute of General Medical Sciences. 


\clearpage


\onecolumngrid
\appendix

\section*{Supplementary materials}

\subsection*{ S1: Cross Sectional Classifier Performance}

      \textbf{FIG. S1} Calibrated Classifier Performance on 1 000 User Geo Tweet Dataset.  Correctly classified humans (True Negatives), are coded in Green, while correctly identified automatons (True Positives) are coded in red.  The 400 tweet average optimal thresholds from the cross validation experiment designate the thresholding for each feature.  The black lines demonstrate each feature cutoff.  

  \includegraphics[scale=.5]{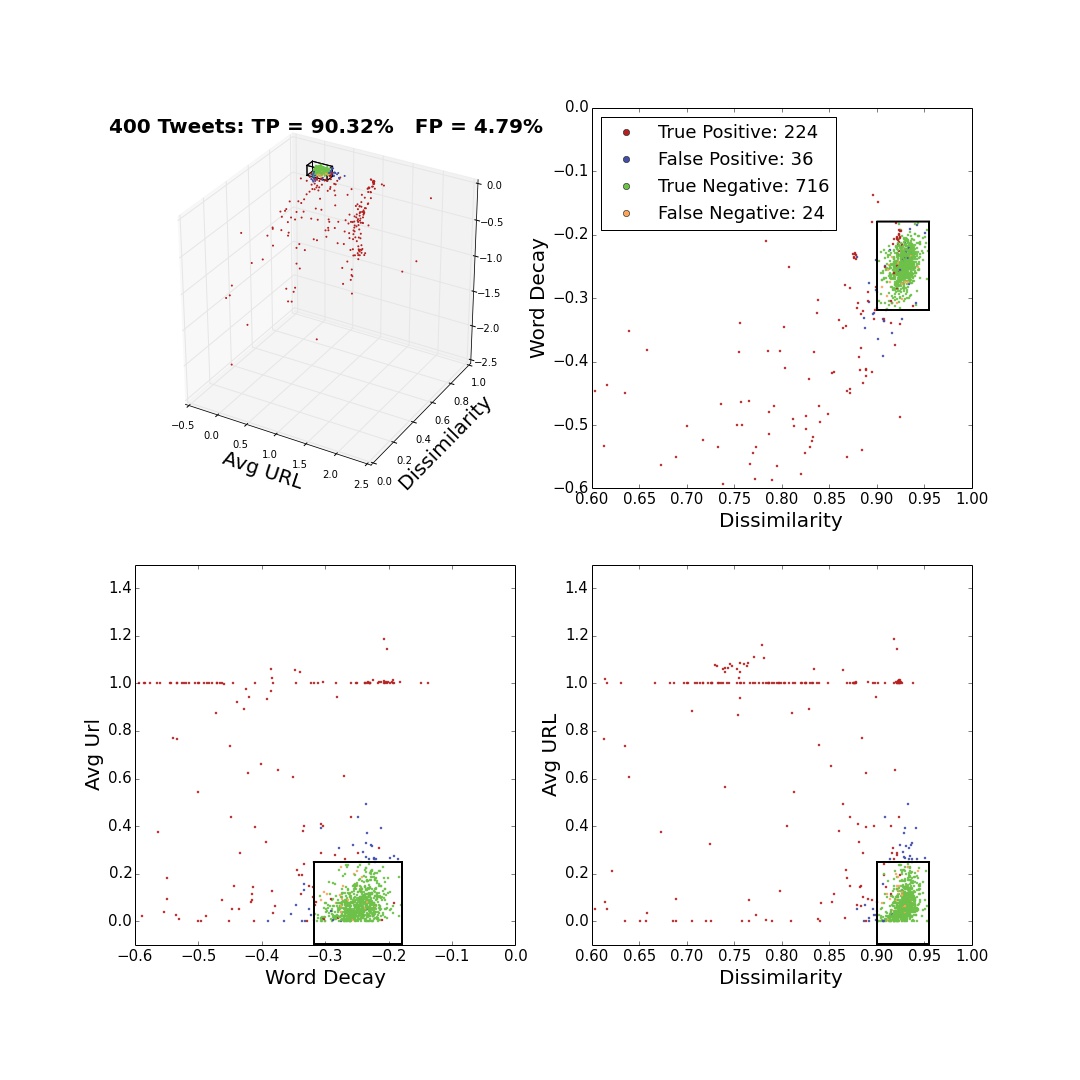}   
 
\pagebreak

\subsection*{ S2: Model Comparisons}

     Receiver Operator Characteristic (ROC) Curves of the performance of each individual feature are given in  \textbf{FIG. S2}  below.  We see each feature set performs comparably with accuracies (measured as AUC) ranging from 80\% to 91\% depending on the number of tweets compiled in the analysis.   Combinations of each metric greatly increases the classification accuracy, the apparent most accurate model  uses all three features.  However, it is notable that combinations of two of these features perform strongly in comparison.   It is also notable that the word introduction decay parameter coupled with the average URL rate performs as well as the Dissimilarity-URL model. The Dissimilarity metric requires determining the Longest Common Substring between many sets of tweets which is computationally expensive compared to analytically calculating the Word Introduction Decay  Parameter.  

\includegraphics[width=.99\linewidth]{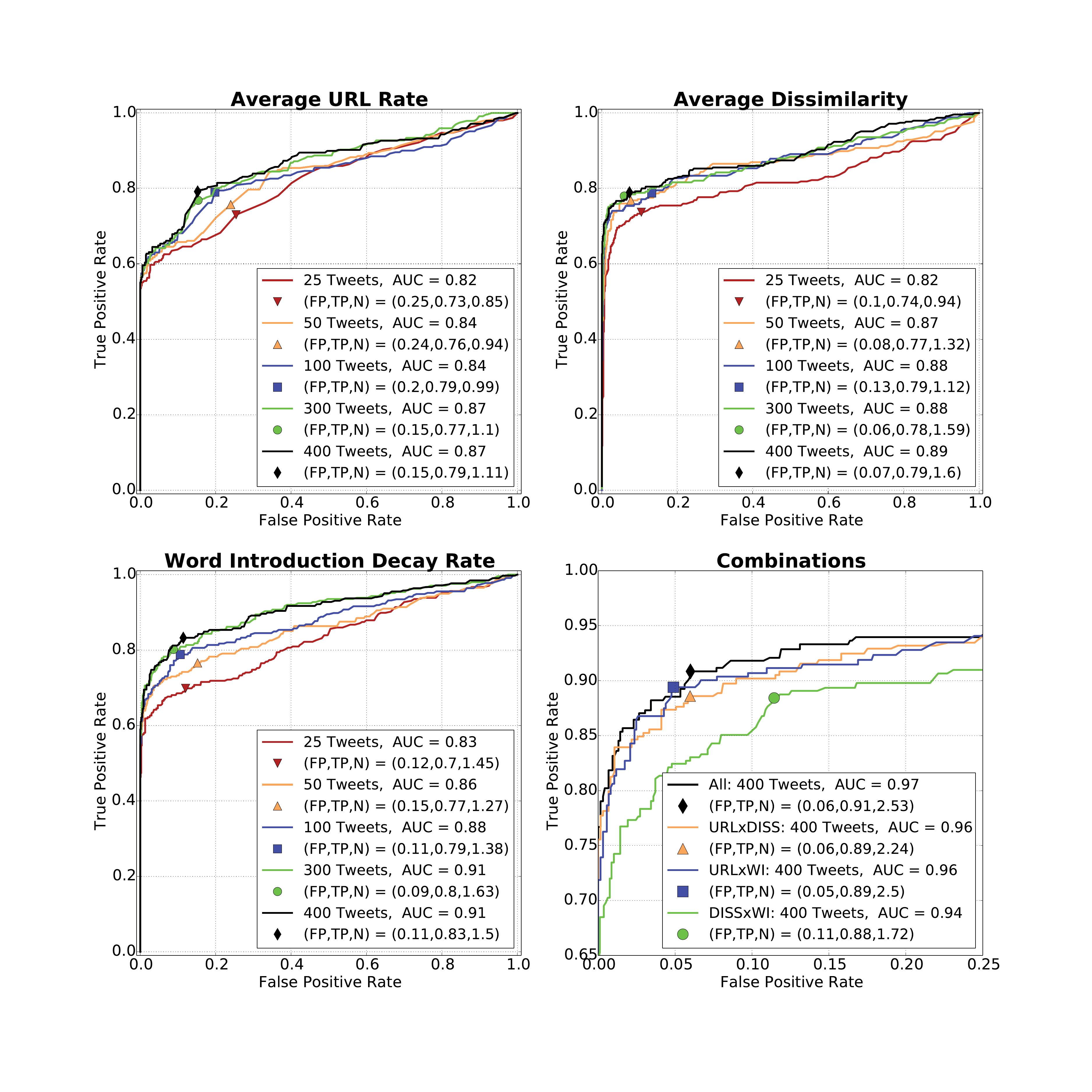}

\pagebreak

\subsection*{S3: Word Introduction Decay Parameter}

  Here, we expand upon our description of the Word Introduction Decay Parameter.  This parameter is based upon random word shufflings of a text, but is computed via the analytic formula given in Eq 8. of [16].  To determine the decay rate parameter, we: (1) compute the word introduction rate as a function of word number, $n$,  and (2) regress in log-log space for a power law decay rate parameter measuring in the final third for the tail, where the decay rate assumes the form of a power law.
While this heuristic is crude and could certainly be refined to more precisely measure the power law region, which can vary with corpus size, the tightness of organic-user clustering afforded by this parameter,
coupled with its computationally cheap cost  when compared to the pairwise tweet dissimilarity metric affords us great power for bot discrimination.

In \textbf{Figure S3} below  the unique word introduction gaps are plotted in log-space as a function of unique word introduction number (rank) for each individual in our data set for various numbers of tweets.  We see the distribution growing with the number of tweets.  However, at each resolution, the human class is very distinctly distributed in comparison to each form of automation.  Even within 25 tweets, the human clustering is visually apparent versus their automated counterparts. 

\includegraphics[width=.99\linewidth]{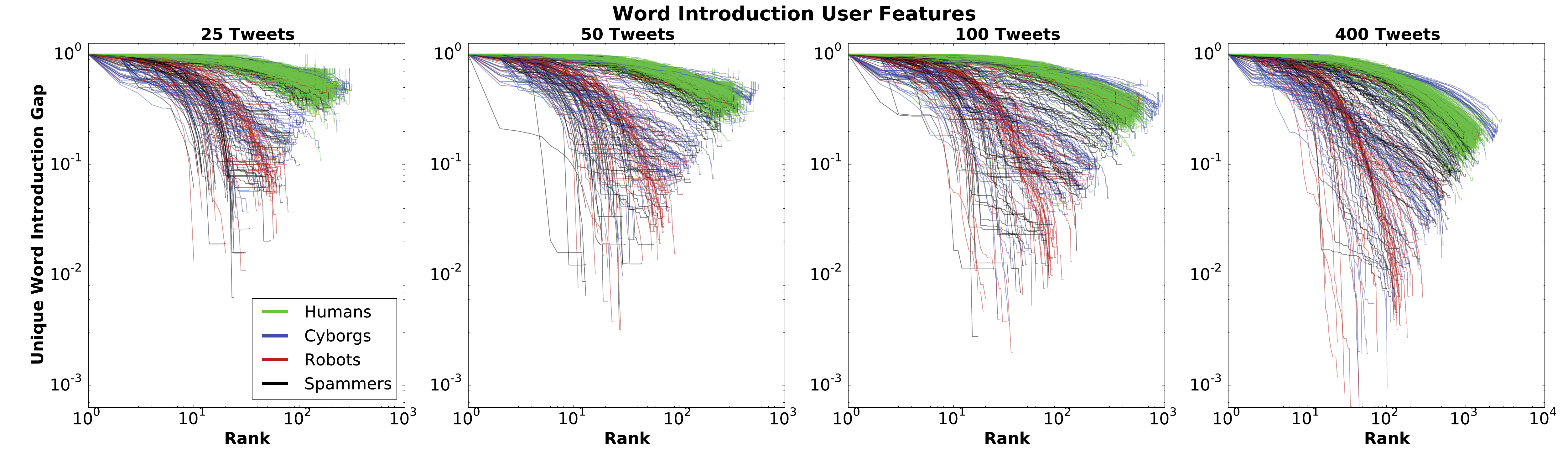}

 In \textbf{ Figure S4} below, we visualize the stability of this parameter between an individual's set of tweets.  The tweets from each account in our data set were resampled 100 times to recompute the word introduction decay parameter for 25 tweets (top) and 400 tweets (bottom).  The standard deviation between each account's 100 decay parameters is given in the histogram.  The average standard deviation across all individuals of each set, $\mu$, is given in the title of each histogram.  Notably, the humans have very little deviation (i.e. within the 'window of forgiveness') for both sets of tweets.  Automated classes, in particular spammers,  can vary quite wildly depending on the sample of tweets that are analyzed.  In particular, spammers look similar to (and usually are) humans and if the spamming event is not captured in the sampled data they will be misclassified.  This decay parameter for human text is robust for varying sets of tweets and is quite distinguishable from automated accounts.

\begin{center}
\includegraphics[width=.99\linewidth]{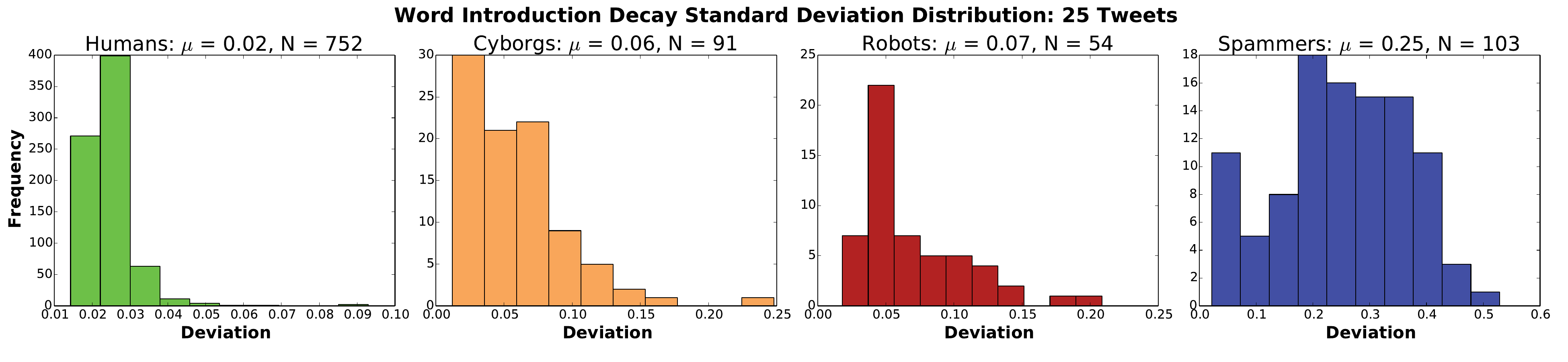} 
\includegraphics[width=.99\linewidth]{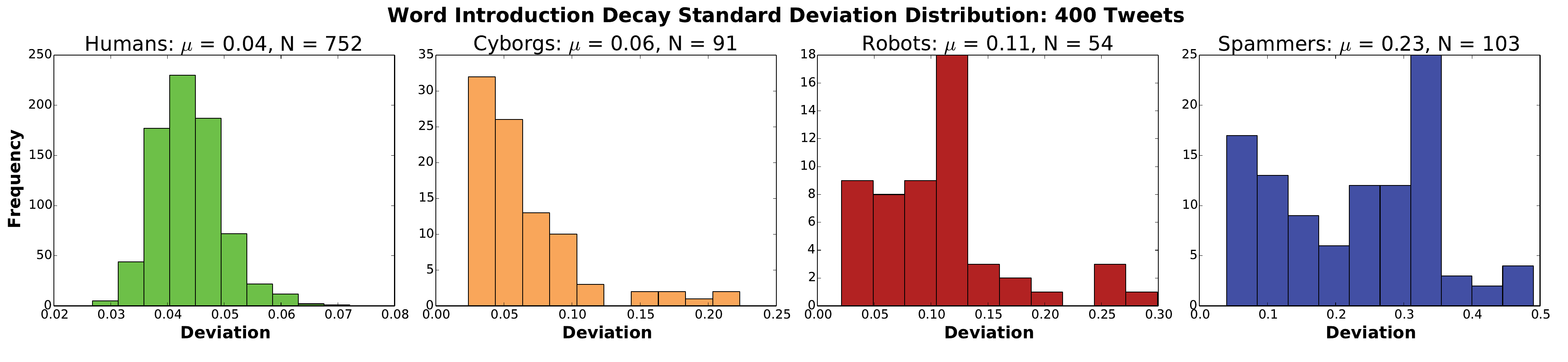}
\end{center}

\goodbreak
\pagebreak

\FloatBarrier
\subsection*{S4: Human Annotation of Twitter Accounts}

\twocolumngrid
  In this section we describe the annotation process for classifying user accounts.     Each of the one-thousand accounts were separately classified by two evaluators.  All of the collected tweets from each account were assessed until the presence of automation was uncovered.  An account was coded as `human' if no automated posting presence was detected (i.e. an algorithm posting on the individual's behalf).  The inter-rater reliability is summarized in the table below by listing the classification discrepancies between account classes.  For each class,  the counts of the type of rating are displayed.  For example,  the Human class had a total of 6 account discrepancies which is composed of 12 different scores (6 from each rater) -  5 human codings, 3 robot, 0 cyborg, and 4 spammers.

 The reliability between raters was favorable (91.49\%) for the entire dataset.  The largest source of discrepancies were from the `Robot' and `Spammer' classes.  The 'Spammer' class was most confused with 'Humans'- which is intuitive because many of these individuals were humans that utilized a algorithm to SPAM a particular message.   `Robots' were commonly confused with 'cyborgs'.  This is most likely due to boundary cases regarding both classes.  The boundaries between these classes can at time be ambiguous.  We classified cyborgs as automatons that were posting `borrowed content` from another source or an account that used human assisted automation, i.e. a human that could be overseeing an automated account.  Robots were defined as strictly posting structured automated messages in the forms of updates.  Perhaps future work can work to sub classify different types of robots and cyborgs to investigate the ecology of these automatons.
 
 \begin{table}[!htbp]%
\caption{Annotation Discrepancies of Twitter Accounts}
\begin{tabular}{|l|cccc|c|}
\cline{1-6}
  \textbf{Class} & &    \textbf{Discrepancies} & &  & \textbf{Totals}  \\
\cline{2-5} 
  & Human & Robot & Cyborg & Spammer &  \\
 \hline
  Human & 5 & 3 &  0 & 4 & 0.79\% \\
  \hline
   Cyborg & 2 & 9 & 6 & 1 & 9.89\% \\
   \hline
  Robot & 0  & 31 & 32 & 1 & 60.38\%  \\
  \hline
  Spammer & 31 & 4 & 4 & 37 & 36.89\% \\
  \hline
  All & 6 & 9 & 32 & 38 & 8.51\% \\
\hline
\end{tabular}
\end{table}

  Each discrepancy was revisited by both annotators and discussed until a class was determined.   For extreme boundary cases, the account ID was searched via the hyperlink:  \url{https://twitter.com/intent/user?user_id=#####}.   This helped observe other user features (screen name, description, etc.)  to make a better decision about the user.  This was especially helpful for identifying promotional accounts or news sources.  


  Screen shots of particular accounts are given below to help describe the annotation process.  Each annotator scrolled through a terminal interface containing each individual's tweets.  Scrolling through 'human` text appears un-ordered and chaotic with very little structure.  Automated accounts have very structured messages, hence these patterns become very apparent in comparison to human accounts.   \\

\textbf{Cyborg Account Example:}
  A canonical cyborg's tweets are given below.  This particular automaton is a news promotional account that is tweeting links to articles.  Notice the description tailors off when it reaches the character limit and shows this with an ellipses (...) next to a URL.  
\begin{center} \includegraphics[width=.99\linewidth]{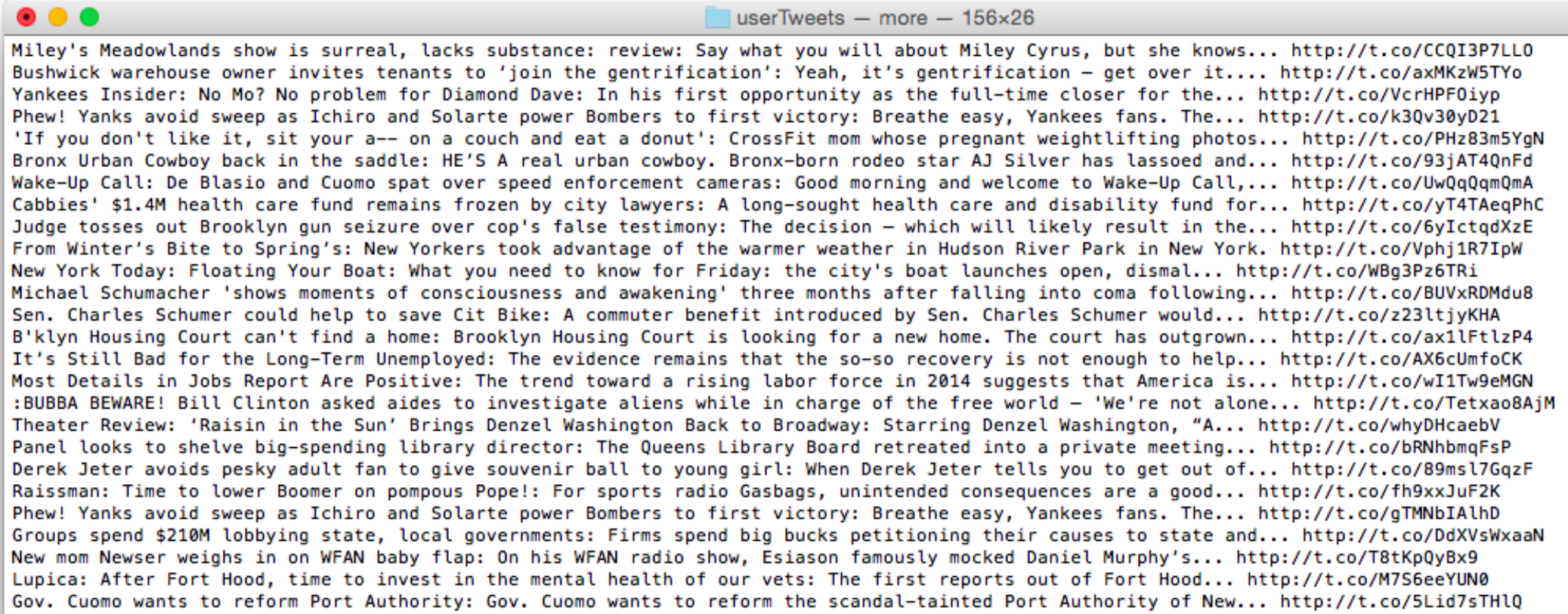} \end{center}

\textbf{Robot Account Example:}
Robots tweet generically structured messages, usually as a form of update.  These automatons have a very limited vocabulary and in general only change a few characters per tweet.  This robot (below) is an example of a weather update bot that is tweeting statistics about the weather at regular intervals.  
\begin{center} \includegraphics[width=.99\linewidth]{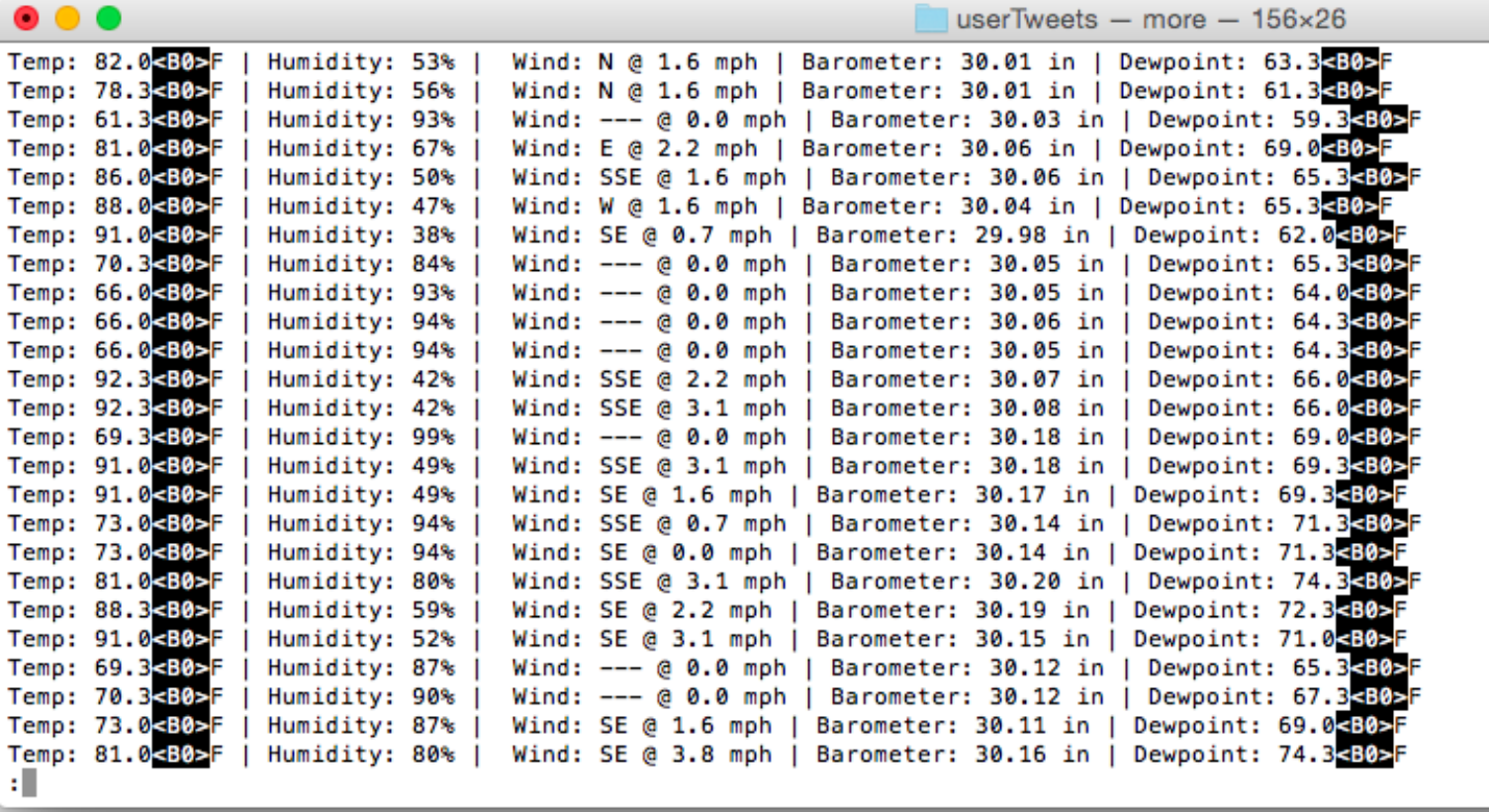}  \end{center}

\textbf{Spammer Account Example:}
  Tweets from a spamming human account are given below.  This individual has utilized an algorithm to tweet at a musical celebrity.  Many of these spam algorithms try to fool Twitter's detector by including a different number or symbol at the end of the tweet.     

\begin{center} \includegraphics[width=.99\linewidth]{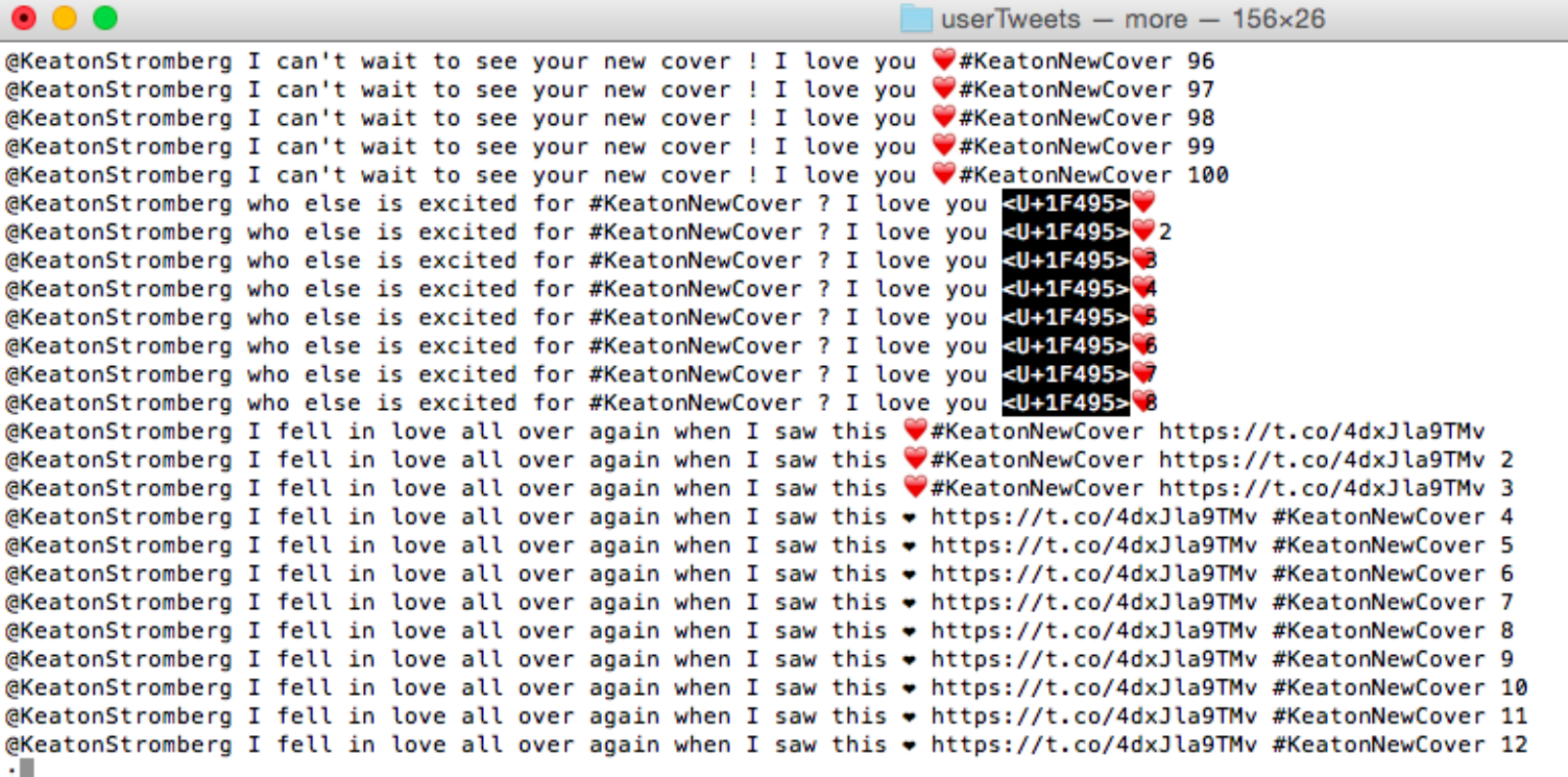} \end{center}

\goodbreak
\pagebreak

\onecolumngrid
\FloatBarrier
\subsection*{ S5: Mixed User Sample Tweets}

\begin{table}[!htbp]
\scriptsize
\caption{ \large{\textbf{Robot Sample Tweets}}} 
\hskip-2.0cm
\begin{tabular}{|l|l|}
\cline{1-2}
No. & Tweet \\ 
\hline
1 & \#SuspiciousPerson 3955 W D JUDGE DR 32808 (7/26 22:19) \#Orlando \#MercyDrive \\
\hline
 2 & \#AccidentWithRoadBlockage LEEVISTA BV \& S SEMORAN BV N/A (6/24 18:07) \#Orlando \#CentralBusinessDistrict \\
\hline
3 & @USER the 1st mention of \#sunnysmiles appears on your TL. Now is Trending Topic in United States! \#trndnl \\
\hline
4 & TRAFFIC STOP at SE 181ST AVE / SE PINE ST, GRESHAM, OR [Gresham Police \#PG14000039852] 23:58 \#pdx911 \\
\hline
5 & A 2002 Ford Ranger was just scanned near Cleveland, TN 37311 URL \#myvinny \#startup \#buyacar \\
\hline
6 & Visiting \#SantaCruz, \#California? Check out this great new app for news, weather, hotels, and food here! URL \\
\hline
7 & Trend Alert: \#heatJustinBieber. More trends at URL \#trndnl URL \\
\hline
8 & Temp: 76.5°F | Humidity: 70\% | Wind: --- @USER 0.0 mph | Barometer: 30.01 in | Dewpoint: 66.0°F \\
\hline
9 & On Sunday 4, \#WinUgly was Trending Topic in Pittsburgh for 10 hours: URL \#trndnl \\
\hline
10 & Wind 0.0 mph ---. Barometer 1016.0 mb, Rising slowly. Temperature 67.6 °F. Rain today 0.46 in. Humidity 96\% \\
\hline
\hline
\end{tabular}
\end{table}

\begin{table}[!htbp]
\scriptsize
\caption{ \large{ \textbf{ Cyborg Sample Tweets}}} 
\hskip-2.0cm
\begin{tabular}{|l|l|}
\cline{1-2}
No. & Tweet \\ 
\hline
1 & Indianapolis, IN suburb Family practice physi... - Soliant Health: (\#Indianapolis, IN) URL \#FamilyPractice \#Job \\
\hline
2 & Barnabas Health: Patient Care Associate (\#LongBranch, NJ) URL \#Nursing \#Job \#Jobs \#TweetMyJobs \\
\hline
3 & Overlake offers a low-cost way to check lungs for cancer early: Doctors at Overlake say they?re tired of waiting... URL \\
\hline
4 & \#TweetMyJobs \#Nursing \#Job alert: Opening | Accountable Healthcare Staffing | \#Glendale, AZ URL \#Jobs \\
\hline
5 & Soliant Health \#IT \#Job: Cerner Jobs - Cerner Analyst - San Diego, CA ( \#SanDiego , CA) URL \#Jobs \#TweetMyJobs \\
\hline
6 & Tyco \#Marketing \#Job: Digital Marketing Specialist ( \#Monroe , NC) URL \#Jobs \#TweetMyJobs \\
\hline
7 & @USER Timing is everything when announcing a breakup URL \\
\hline
8 & Southwest flights briefly diverted to DFW Airport on Friday: A Southwest Airlines plane experiencing mechanical... URL \\
\hline
9 & Fort Carson To Welcome Home About 225 Soldiers: FORT CARSON, Colo. (AP) ? Fort Carson will welcome home about 225... URL \\
\hline
10 & Joint venture secures \$97M in financing for two Boston hotels: Commonwealth Ventures and Ares Management have... URL              \\

\hline
\hline
\end{tabular}
\end{table}

\begin{table}[!htbp]
\scriptsize
\caption{ \large{\textbf{Spammer Sample Tweets}}} 
\hskip-2.0cm
\begin{tabular}{|l|l|}
\cline{1-2}
No. & Tweet \\ 
\hline
\hline

1 & \#CallMeCam n\#CallMeCam n @USER n nIf Cameron called me it'll seriously make my day I love you please call me! 100 \\
\hline
2 & \#CallMeCam n\#CallMeCam n @USER n nIf Cameron called me it'll seriously make my day I love you please call me! 321 \\
\hline
3 & \#CallMeCam n\#CallMeCam n @USER n nIf Cameron called me it'll seriously make my day I love you please call me! 167 \\
\hline
4 & S/o to @USER thanks for the support. Check out my music @USER URL I promise u won't be disappointed. \\
\hline
5 & S/o to @USER destiiny thanks for the support. Check out my music @USER URL I promise u won't be disappointed. \\
\hline
6 & S/o to @USER thanks for the support. Check out my music @USER URL I promise u won't be disappointed. \\
\hline
7 & nAshton Irwin from 5SOS n nMy birthday is in 11 days, nAnd it would be an amazing gift, nIf you could follow me. Ily n @USER n nX3126 \\
\hline
8 & nAshton Irwin from 5SOS n nMy birthday is today, nAnd it would be an amazing gift, nIf you could follow me. Ily n @USER n nX5408 \\
\hline
9 & nAshton Irwin from 5SOS n nMy birthday is in 22 days, nAnd it would be an amazing gift, nIf you could follow me. Ily n @USER n nX765 \\
\hline
10 & nAshton Irwin from 5SOS n nMy birthday is in 8 days, nAnd it would be an amazing gift, nIf you could follow me. Ily n @USER n nX3422 \\

\hline
\hline
\end{tabular}
\end{table}

\begin{table}[!htbp]
\scriptsize
\caption{ \large{\textbf{Human Sample Tweets}}} 
\hskip-2.0cm
\begin{tabular}{|l|l|}
\cline{1-2}
No. & Tweet \\ 
\hline
\hline

1 & I'll marry whoever comes thru with some food \\
\hline
2 & Ewwww them seats \#BlackInkCrew \\
\hline
3 & @USER if you only knew ???? i like you \\
\hline
4 & Really wish he wasn't so **** busy ?? \\
\hline
5 & My son's name is Gabriel. \\
\hline
6 & guess I need to get up and get ready then \\
\hline
7 & Grandma stayed on me bout not wearing socks in her house aint nobody got time for that \\
\hline
8 & Thank you for reading. ?? \\
\hline
9 & @USER: If only I knew.. ??? \\
\hline
10 & WHY ARE YOU SO HOT URL \\

\hline
\hline
\end{tabular}
\end{table}

\end{document}